\pdfoutput=1

\documentclass[11pt]{article}

\usepackage[final]{acl}

\usepackage{times}
\usepackage{latexsym}
 
\usepackage[T1]{fontenc}

\usepackage[utf8]{inputenc}

\usepackage{microtype}

\usepackage{inconsolata}

\usepackage{graphicx}
\usepackage{multirow}
\usepackage{amsmath}
\usepackage{amssymb}
\usepackage{booktabs}
\usepackage{subcaption}
\usepackage{bm}
\usepackage{tcolorbox}
\usepackage{enumitem}
\usepackage{svg}

\usepackage{cleveref}
\usepackage{makecell}

\usepackage{color}

\usepackage{listings}
\title{Think in Sentences: Explicit Sentence Boundaries Enhance Language Model's Capabilities}

\author{
    Zhichen Liu, 
    Yongyuan Li, 
    Yang Xu\textsuperscript{*}\\
    Department of Computer Science and Engineering\\ Southern University of Science and Technology\\
    \href{mailto:liuzc2024@mail.sustech.edu.cn}{liuzc2024@mail.sustech.edu.cn}, \href{mailto:xuyang@sustech.edu.cn}{xuyang@sustech.edu.cn}\\
}

\begin{document}
\maketitle

\insert\footins{
  \footnotesize 
 $^*$ Corresponding author.
}

\begin{abstract}

Researchers have explored different ways to improve large language models (LLMs)' capabilities via dummy token insertion in contexts. However, existing works focus solely on the dummy tokens themselves, but fail to leverage the inherent sentence-level structure of natural language. 
This is a critical oversight, as LLMs acquire linguistic capabilities through exposure to human-generated texts, which are inherently structured at the sentence level.
Motivated by this gap, we propose an approach that inserts delimiters at sentence boundaries in LLM inputs, which not only integrates dummy tokens into the context, but also facilitates LLMs with sentence-by-sentence processing behavior during reasoning.
Two concrete methods: (1). In-context learning and (2). Supervised fine-tuning are experimented using 7B models to 600B Deepseek-V3.
Our results demonstrate consistent improvements across various tasks, with notable gains of up to 7.7\% on GSM8k and 12.5\% on DROP. Furthermore, the fine-tuned LLMs can incorporate sentence awareness evidenced by their internal representations.
Our work establishes a simple yet effective technique\footnote{A demonstrative code repository is provided: \url{https://github.com/CLCS-SUSTech/think-in-sentence}. 
} for enhancing LLM's capabilities, offering promising directions for cognitive-inspired LLM enhancement paradigm.

\end{abstract}

\section{Introduction}
Sentence-level structure has long been a cornerstone of early neural language models: Skip-thought vectors \citep{skip} were trained to reconstruct neighboring sentences, while BERT’s next-sentence prediction task \citep{bert} proved indispensable for downstream performance by encoding inter-sentence coherence. 
Yet with the rise of large language models (LLMs), whose success stems primarily from scaling pretraining on massive unstructured text, sentence boundaries have been increasingly sidelined, treated as indistinguishable from ordinary tokens in the token-by-token processing pipeline. This oversight is striking: human language generation relies on incremental, sentence-by-sentence cognition, but LLMs learn from the continuous text that results from this process, creating an inherent misalignment between human cognitive mechanisms and model input processing.

\begin{figure*}[t]
    \centering
    \includegraphics[width=0.9\linewidth]{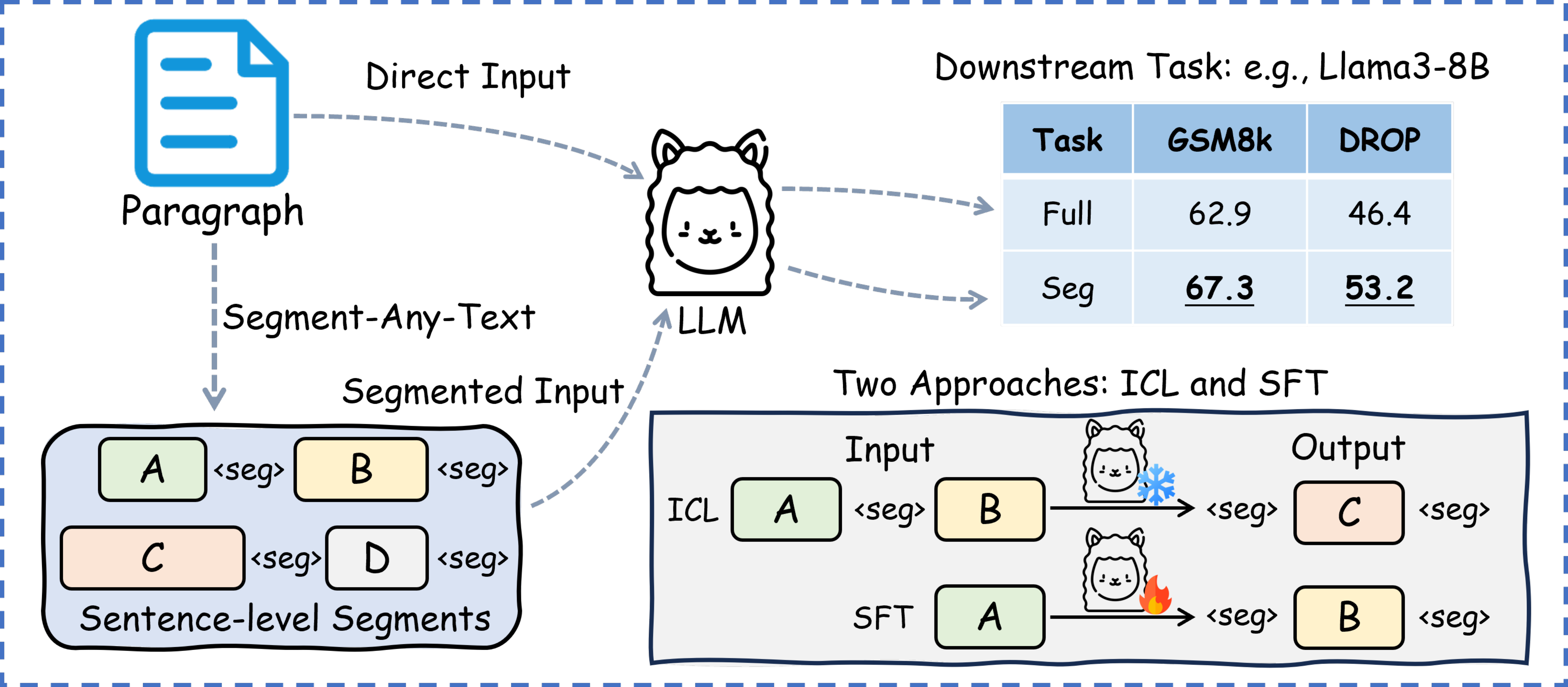}
    \caption{\textbf{Overview of Sentence-Level Inference:} We insert delimiters at sentence boundaries to enable LLMs to ``pause and integrate context'' during inference. Two approaches are proposed: (1) In-Context Learning (ICL): LLMs infer with delimiter placement from exemplars in long contexts; (2) Supervised Fine-Tuning (SFT): LLMs learn sentence-segmented patterns via delimiter-inserted training data. For Llama3-8B-Instruct, this approach improves performance by $\sim$4.4\% on GSM8k and $\sim$6.8\% on DROP over unsegmented inputs.}
    \label{fig:workflow}
\end{figure*}

Against this backdrop, we argue that re-emphasizing sentence-level information offers a largely untapped avenue to enhance LLMs, especially for ``free-lunch'' (cost-neutral) improvements. Since GPT series \citep{gpt2,instructgpt} established modern LLM training paradigms, efforts to improve performance have followed two main paths: training-time scaling (e.g., scaling laws for model/data size \citep{oaiscalinglaw, chinchillascalinglaw, palm, llama}) and test-time scaling (e.g., instruction for thinking step-by-step \citep{cot,tot}, or reinforcement learning (RL) for self-reflection \citep{reflection, rstar, restmcts}). However, these methods incur substantial costs: training-time scaling demands massive compute or data, while test-time scaling increases inference latency and token consumption.

To address this, recent work \citep{thinkspeaktraininglanguage} proposed inserting special ``pause'' tokens into contexts as a free-lunch alternative, obtaining performance gains without extra costs. Yet this approach suffers from limited robustness and generality: dummy token placement lacks linguistic priors, requiring manual tuning across tasks and does not leverage the inherent structure of human language. This gap raises our research question: \textbf{Can we design an effective strategy that harnesses sentence-level linguistic priors to robustly enhance LLM performance?}

\subsection{Main Contributions}

We introduce a \textit{sentence-level inference} paradigm that accentuates sentence boundaries via task-agnostic delimiters, bridging the gap between LLMs’ token-by-token processing and the more human-like sentence-by-sentence cognition process. Our key contributions are threefold:

\paragraph{Paradigm Innovation:} Unlike explicit reasoning prompts (e.g., CoT), we implicitly enhance inference by inserting delimiters at sentence boundaries. These delimiters act as ``inference anchors'' -- not mere grammatical markers -- to trigger a ``context integration $\rightarrow$ next-step planning'' cycle at the end of each sentence, thereby simulating human post-sentence reflection.

\paragraph{Dual Implementation:} We propose two complementary methods to instantiate this paradigm: (a) ICL, where LLMs learn delimiter placement from contextual exemplars (suited for long-input scenarios); (b) SFT, where models are fine-tuned on delimiter-inserted, sentence-segmented data (for short-input tasks). Both methods require minimal overhead, qualifying as free-lunch strategies.

\paragraph{Empirical and Mechanistic Insights:} Across model scales (7B to 600B), our methods yield consistent downstream gains (e.g., $\sim$7.7\% on GSM8k, $\sim$12.5\% on DROP). Ablations reveal that: (i) structured delimiters outperform arbitrary tokens for ICL; (ii) sentence-level segmentation is the optimal granularity; (iii) gains arise from synergy between LLMs’ Chain-of-Thought reasoning and sentence-level inference. We further validate mechanisms via attention map visualization, showing that delimiters capture more information than normal tokens.

\subsection{Related Works}

\paragraph{Test-Time Scaling for LLMs}
Test-time scaling aims to improve performance by extending inference ``thinking time.'' CoT \citep{cot} and ToT \citep{tot} use instruction prompts to elicit step-by-step reasoning, while follow-ups add self-verification \citep{reflection} or RL-driven search (e.g., MCTS \citep{rstar, restmcts}) to explore solution spaces. RL has also been applied to training (e.g., DeepSeek R1 \citep{r1}, Kimi K1.5 \citep{k1.5}) to teach self-exploration. While effective, these methods drastically increase inference latency and token costs, limiting deployment.

\paragraph{Pause/Dummy Token Strategies}
\citet{thinkspeaktraininglanguage} pioneered cost-neutral test-time scaling via inserting pause tokens, showing gains in pretraining/fine-tuning for 1B-scale models. However, their approach has critical limitations: (i) no validation on large-scale LLMs ($\geq$7B parameters); (ii) token count requires task-specific manual tuning; (iii) lack of linguistic priors leads to limited robustness across tasks.

\paragraph{Sentence-Level Granularity in LLMs}
Recent work has revisited sentence-level structure for LLMs, though with different goals. \citet{sentreward} proposed a sentence-level reward model that outperforms token/response-level alternatives for alignment. \citet{gspo} replaced GRPO’s \citep{grpo} token-level objective with sequence-level optimization, improving stability. These works validate the value of sentence-level paradigm in their objectives, while our work targets inference-time, free-lunch performance gains via sentence-level inference.

Beyond sentence-level boundaries, recent studies have also explored incorporating finer-grained syntactic and semantic structures into prompt engineering. For instance, leveraging syntax trees has shown benefits in specific structured tasks like aspect-based sentiment analysis \cite{labate2024infusingpromptssyntaxsemantics} and semantic infusing \cite{related2}; however, extending such complex syntactic augmentations to general-purpose reasoning scenarios remains an open and promising direction.

\section{Method}\label{sec:2}

Our central hypothesis is that by explicitly modeling sentence boundaries, we can induce a more structured, sentence-by-sentence reasoning process in LLMs, thereby enhancing their performance on complex downstream tasks. To this end, we reformulate the standard language modeling objective to incorporate sentence-level structural information. We introduce a special delimiter token, denoted as ``$x_{seg}$'', which is inserted at the end of each sentence. This transforms a text sequence $T$:
\begin{equation}
    T=[t_1,t_2,t_3,...,t_n]
\end{equation}
into a structurally-annotated sequence $S$:
\begin{equation}
    S=[s_1,x_{seg},s_2,x_{seg},...,s_n,x_{seg}]
\end{equation}
Here, each $s_i$ represents a sentence from the original text $T$, consisting of multiple tokens $t$. Consequently, the model's objective is no longer limited to predicting the next token in a flat sequence; it further entails learning the optimal timing to generate the delimiter ``$x_{seg}$''. In doing so, the model performs implicit sentence segmentation as part of its generative objective. Despite simplicity, this modification effectively encourages the model to better recognize and leverage sentence-level semantics. We explore two primary strategies to implement this capability in LLMs: In-Context Learning and Supervised Fine-Tuning.

\subsection{Sentence-Aware Prompting via In-Context Learning}

In-Context Learning (ICL) offers a lightweight, inference-time approach to elicit desired behaviors from LLMs without updating the model weights. We use ICL to guide the model to adopt a sentence-delimited generation style. This is achieved by including few-shot examples in the prompt, where each sentence within the demonstration is explicitly terminated by the predefined delimiter. 
The model is then tasked with completing the final, incomplete example. 
The generation process follows the standard auto-regressive objective, but the context primes the model to continue the observed pattern:
\begin{equation}
y_t=\mathop{\text{argmax}}\limits_{y}\ P(y|C_{\text{few-shot}},Q,y_{<t};\theta) \label{eq:2}
\end{equation}
where $C_{\text{few-shot}}$ is the context containing sentence-delimited examples, $Q$ is the user's query, and $\theta$ represents the frozen model parameters. 
According to \citet{in-context}, the model learns from analogy to structure the intermediate reasoning and the output in a sentence-by-sentence manner. 
As validated in experiments in \Cref{sec:3}, this ICL-enabled structured generation process leads to stable performance gains. 
However, the efficacy of ICL is contingent on the availability of sufficient context length for demonstrations, limiting its applicability in zero-shot or context-constrained scenarios.

\subsection{Internalizing Sentence Structure via Supervised Fine-Tuning}\label{sec:sft}

To overcome the limitations of ICL and to build a more robust, inherently sentence-aware model, we propose a Supervised Fine-Tuning (SFT) strategy. This approach aims to internalize the sentence-level structural prior directly into the model's parameters, making the behavior more \emph{intrinsic} rather than context-dependent.

First, we curate a fine-tuning dataset by systematically preprocessing a collection of large-scale text corpora, with delimiters inserted at every sentence boundary. Then we fine-tune the language model on this modified dataset using the standard causal language modeling (CLM) objective. The loss function is rewritten to reflect the sentence-level training objective as follows:
\begin{equation}
\begin{aligned}
    &\mathcal{L}_{SFT}(\theta)=\sum^S_{s'\in S}\sum_{i=1}^{|s'|}\log P(t_i|t_{<i};\theta)\\
    &\text{where $s'=[s,x_{seg}]$ and $t_{|s'|} = x_{seg}$}
\end{aligned}
\end{equation}
Through the training process, the model learns to predict sentence boundaries, which it integrates as a fundamental component of language generation. 
For implementation, we add the delimiter as a special token into the tokenizer, thereby introducing new embeddings and LM head weights. Compared to ICL, the SFT approach yields a model that natively generates sentence-delimited text, making it more effective for zero-shot applications and better aligned with real-world deployment scenarios where concise prompts are preferred. 

\section{Experiments}\label{sec:3}

We conduct a comprehensive suite of experiments to validate our central hypothesis: inducing sentence-level awareness in LLMs enhances their reasoning capabilities. We aim to answer two concrete research questions:
\begin{enumerate}[leftmargin=1.2em]
\item \textbf{RQ1:} Does prompting with sentence delimiters during inference (i.e., the ICL approach) improve performance on reasoning tasks across various model scales?
\item \textbf{RQ2:} Can such sentence-aware behavior be permanently internalized via fine-tuning (i.e., the SFT approach), and how does this compare to standard fine-tuning and other methods? 
\end{enumerate}

\subsection{Experiment Setup}

\paragraph{Models.} Our experiments span various sizes of LLMs. For ICL, we evaluate open-source LLMs including \textbf{LLaMA3-8B-Instruct} \cite{llama3} and \textbf{Qwen2-7B-Instruct} \cite{qwen2}, a larger LLM \textbf{Qwen2.5-72B-Instruct} \cite{qwen2.5}, and a SOTA LLM, \textbf{DeepSeek-V3} \cite{dsv3}, via its API\footnote{https://api-docs.deepseek.com/}. For SFT, we perform full-parameter fine-tuning on \textbf{LLaMA3-8B-Base} using 8$\times$NVIDIA L40 GPUs.

\begin{figure}[tbp]
    \centering
    \includegraphics[width=\linewidth]{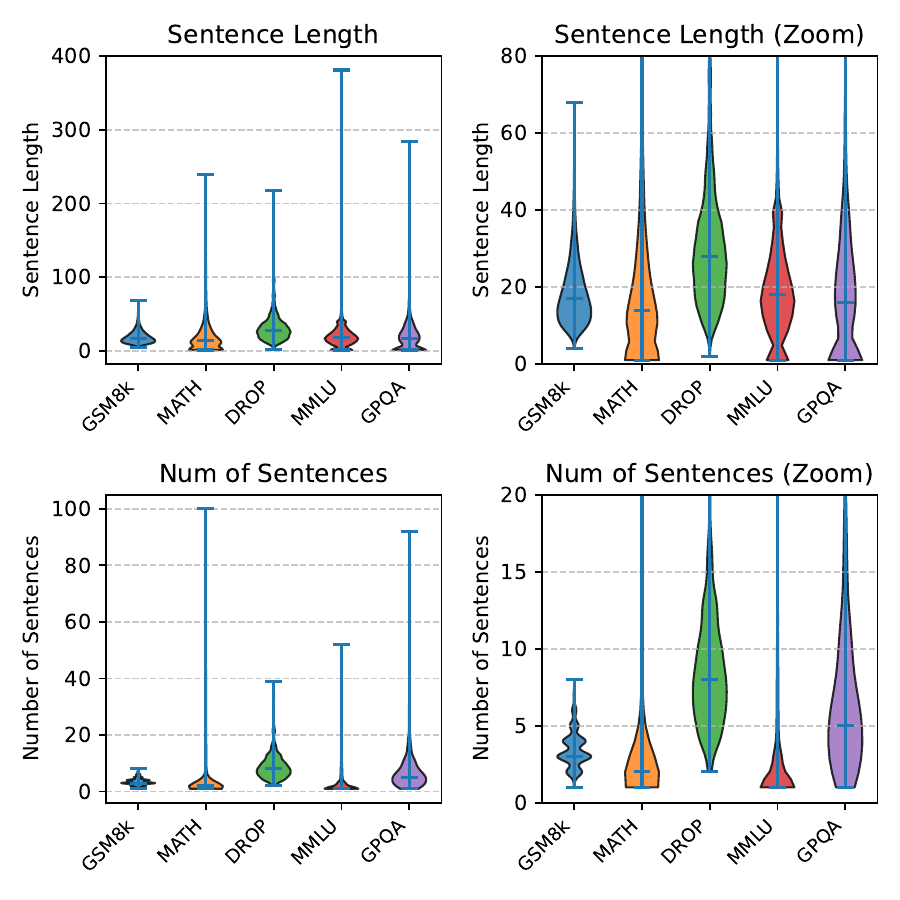}
    \caption{The distributions of sentence lengths and number of sentences for each dataset. The left column figures are the origin distribution, and the right column figures are zoomed-in views. Horizontal bars indicate medians and extrema. Sentence lengths are counted by number of tokens, from the Llama3 tokenizer.}
    \label{fig:dataset-info}
\end{figure}

\paragraph{Datasets and Tasks.} We use a diverse suite of benchmarks targeting on different reasoning types:
\begin{itemize}[leftmargin=1em, itemsep=-0.1em]
\item \textbf{Mathematical Reasoning:} GSM8k \cite{gsm8k} and MATH \cite{math}.
\item \textbf{Reading Comprehension:} DROP \cite{drop}, which requires reasoning over paragraphs.
\item \textbf{General Knowledge Understanding:} MMLU \cite{mmlu} and its more challenging successor, MMLU-Pro \cite{mmlupro}.
\item \textbf{Expert-Level QA:} GPQA \cite{gpqa}, a dataset of graduate-level questions.
\item \textbf{Code Generation:} HumanEval \cite{humaneval} for Python code synthesis.
\end{itemize}
For SFT, we use a curated subset of the TULU3 dataset \cite{tulu3}, from which we exclude safety, multilingual, and table-related data, to focus on general instruction following. \Cref{fig:dataset-info} shows a statistical overview of sentence counts and lengths for the five datasets.

\paragraph{Implementation Details.}
For the purpose of identifying sentence boundaries, we apply the \textsc{SaT-12L-sm} model \cite{sat}, a state-of-the-art sentence segmentation tool, to preprocess all text data, which return sentence boundaries as token positions. Detailed usage see Appendix \ref{app:sat}.  
Then we insert the delimiter token ``$x_{seg}$'' at these boundaries. For SFT, delimiter is added as a new token to the tokenizer, whose corresponding embeddings are learned during training. The evaluation protocols, including few-shot settings for Chain-of-Thought (CoT) prompting, are detailed in \Cref{app:fewshot}. 
Unless otherwise specified, all results are reported using exact match accuracy, with Pass@1 for HumanEval. To ensure a fair comparison, chat templates are disabled for all local evaluations.

\begin{table*}[htbp]
    \centering
    \resizebox{\linewidth}{!}{
    \begin{tabular}{c|ccc|ccc|ccc|ccc}
\toprule
 & \multicolumn{3}{c|}{\textbf{Qwen2-7B-Inst}} & \multicolumn{3}{c|}{\textbf{Llama3-8B-Inst}} & \multicolumn{3}{c|}{\textbf{Qwen2.5-72B-Inst}} & \multicolumn{3}{c}{\textbf{Deepseek-V3}} \\
Dataset & base & seg & $\Delta$ & base & seg & $\Delta$ & base & seg & $\Delta$ & base & seg & $\Delta$ \\
\midrule
MMLU & 64.43 & 69.96 & +5.53\%$\uparrow$ & 62.89 & 67.28 & +4.39\%$\uparrow$ & 86.64 & 86.40 & -0.24\%$\downarrow$ & 74.04 & 74.82 & +0.78\%$\uparrow$ \\
GSM8k & 73.92 & 81.65 & +7.73\%$\uparrow$ & 75.51 & 78.01 & +2.5\%$\uparrow$ & 90.14 & 91.96 & +1.82\%$\uparrow$ & 95.00 & 95.30 & +0.3\%$\uparrow$  \\
MATH & 53.33 & 54.30 & +0.97\%$\uparrow$ & 32.60 & 32.26 & -0.34\%$\downarrow$ & 73.04 & 75.78 & +2.74\%$\uparrow$ & 89.40 & 90.60 & +1.2\%$\uparrow$ \\
DROP & 38.14 & 50.64 & +12.50\%$\uparrow$ & 46.39 & 53.16 & +6.77\%$\uparrow$ & 58.74 & 60.38 & +1.64\%$\uparrow$ & 75.10 & 79.10 & +4\%$\uparrow$ \\
\bottomrule
\end{tabular}
}
    \caption{In-Context Learning results. We compare the performance of vanilla inference (base) against ICL (seg), delimiter here is ``<seg>''. $\Delta$ denotes the absolute improvement. Our method yields consistent gains across models and tasks, with particularly strong improvements on smaller models and in reading comprehension task.}
    \label{tab:icl-main}
\end{table*}
\begin{table*}[htbp]
    \centering
    \resizebox{0.8\linewidth}{!}{
    \begin{tabular}{l|ccccccc}
    \toprule
    & MMLU & GSM8k & MATH & DROP & MMLU-pro & GPQA & HumanEval \\
    \midrule
Std-FT & 59.02 & 72.48 & 30.86 & 48.50 & 34.25 & 26.93 & 56.71 \\
\midrule
Pause-FT & 56.11 & \underline{\textbf{75.44}} & \underline{\textbf{33.50}} & \underline{\textbf{55.97}} & \underline{35.71} & 24.16 & - \\
Seg-FT & \underline{\textbf{60.13}} & \underline{74.91} & \underline{31.58} & \underline{54.26} & \underline{\textbf{40.71}} & \underline{\textbf{27.43}} & \underline{\textbf{62.80}} \\
\bottomrule
    \end{tabular}
    }
    \caption{Supervised Fine-Tuning results on LLaMA3-8B-Base. Our method (Seg-FT) is compared against standard fine-tuning (Std-FT) and pause-token fine-tuning (Pause-FT). Best performance is in \textbf{bold}, and results outperforming the Std-FT baseline are \underline{underlined}. Our approach demonstrates superior robustness and generalization.}
    \label{tab:sft_main}
\end{table*}

\paragraph{Baselines.}
For ICL, the main baseline is the vanilla performance of each model without inserting delimiters. 
For SFT, our method is to fine-tune a Llama3-8B-Base model on the curated TULU3 dataset with delimiters inserted, which we indicated \textbf{Seg-FT}. It is compared with two baselines: Std-FT, a standard fine-tuning baseline, which fine-tunes the same model on the original TULU3 subset \emph{without} inserting delimiters; Pause-FT, a pause-token fine-tuning baseline, which fine-tunes the same model following the settings of \texttt{StdPT\_PauseFT} in \citet{thinkspeaktraininglanguage}, with 10 pause tokens inserted in both training and inference stage. 

\subsection{Results Analysis}

\subsubsection{RQ1: ICL Boosts Reasoning}

As shown in Table \ref{tab:icl-main}, inference with sentence-delimited prompts consistently improves performances across nearly all configurations.

\paragraph{Key Observation 1: Smaller models benefit disproportionately.} The 7B-level LLMs (Qwen2-7B, LLaMA3-8B) exhibit the most significant gains, such as a +7.73\% on GSM8k for Qwen2-7B and +5.53\% on MMLU. This suggests that explicit structural guidance is particularly effective for LLMs with less capacity, helping them organize their reasoning process more effectively. For larger, more capable LLMs (such as Qwen2.5-72B and DeepSeek-V3), the improvements are more modest but still present (smaller in MMLU but larger in MATH and DROP), indicating that even powerful LLMs can benefit from our sentence delimiters-inserted prompting.

\paragraph{Key Observation 2: Performance gains correlate with task types.} The most dramatic improvement is observed on DROP (+12.5\% for Qwen2-7B), a reading comprehension task that requires tracking information across multiple sentences within a context. A reasonable explanation is that by explicitly segmenting sentences, it enable the LLM to process individual facts encoded in separate sentences more effectively, and better understand their relationships, which is important for this type of task.

\begin{figure*}[tbp]
    \centering
    \includegraphics[width=\linewidth]{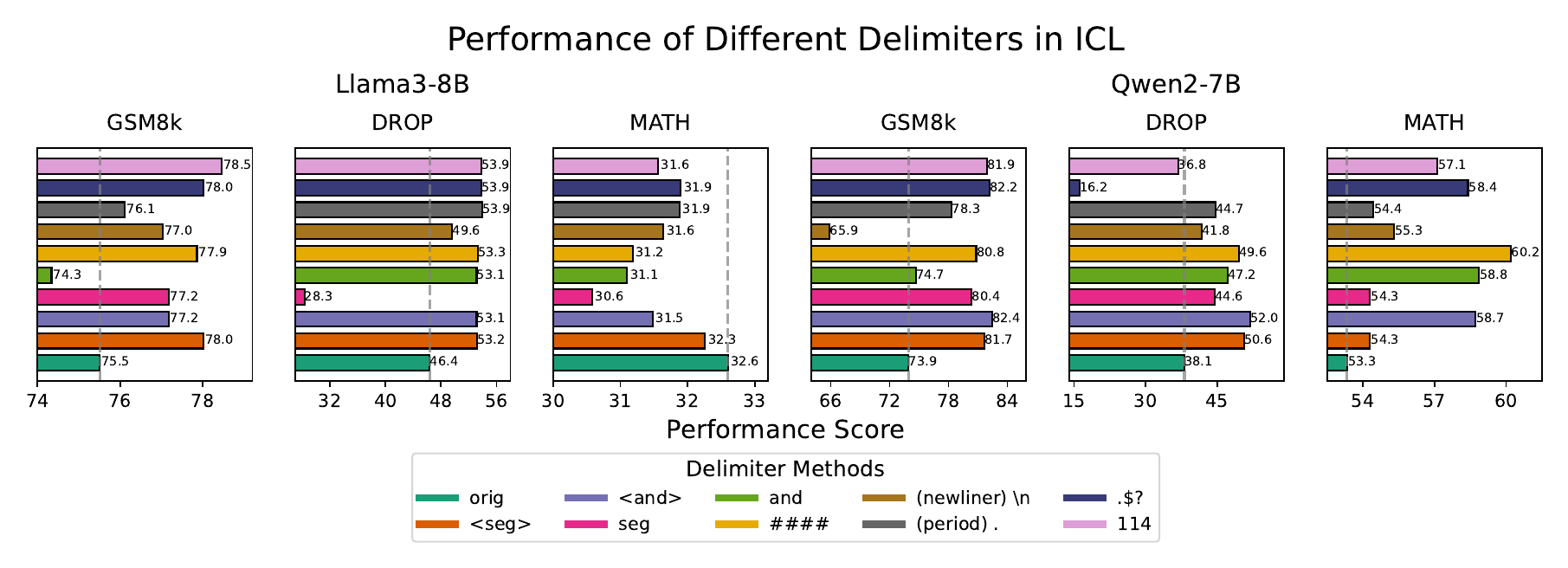}
    \caption{Performance of different delimiter choices in ICL across three datasets. More structured delimiters could consistently yield a better performance, demonstrating the value of a clear, non-semantic structural signal. ``orig.'' denotes the baseline without any delimiters.}
    \label{fig:diff-delimiter}
\end{figure*}

\subsubsection{RQ2: SFT Internalizes Robust Sentence Awareness}
Table \ref{tab:sft_main} shows the results of the SFT approach, yielding several interesting insights. Our method (Seg-FT) has overall better performance than the baselines (Std-FT and Pause-FT). 

\paragraph{Key Observation 3: Sentence-based SFT is more robust than pause-based SFT.} Our method (Seg-FT) consistently outperforms the Std-FT baseline across all seven benchmarks. In contrast, Pause-FT, while staying strong on procedural tasks like GSM8k and MATH, suffers from performance degradation in knowledge-intensive QA tasks like MMLU and GPQA. 
This suggests that while simply ``pausing'' can aid methodical computation, it may disrupt the model's access to or reasoning over its stored knowledge. 
Our method, by encapsulating the generation process into meaningful linguistic units (sentences), seems to provide a more robust and universally beneficial structural prior. 

\paragraph{Surprising Observation: Sentence awareness generalizes to code.} A striking result is the +6.09\% absolute improvement on HumanEval. During inference, we observed that the Seg-FT model is able to insert delimiters within codes. As there exhibits some similar patterns between human language and python code, for example, using newliner as delimiters, it enables the model to learn from the commonalities between the two, thereby acquiring the ability to generalize the segmentation of natural language to code.

\section{Ablation Studies and Analysis}
To analyze what factors contribute to our method's success, we conduct a series of targeted ablation studies. These experiments are designed to answer three fundamental questions: (1) What properties make an effective delimiter? (2) Is sentence-level segmentation truly the optimal strategy for placing these delimiters? (3) What are the underlying mechanisms of delimiters enhancing model performances?

\subsection{On the Importance of a Clear Structural Signal: Delimiter Choice}

In general, we find that the choice of delimiter is non-trivial, and its form and semantics can influence how the model interprets it. We hypothesize that an ideal delimiter should function as a pure structural marker, which is irrelevant of the semantic content of the text. To test this hypothesis, we evaluate a spectrum of delimiters under the ICL setting:
syntactically distinct tokens [``<seg>'', ``<and>'', ``\#\#\#\#''] (structured), common words [``seg'', ``and''] (semantic), punctuation used in human text [``\textbackslash n'', ``.''] (delimiters in natural language), a numeric token [``114''] and a meaningless symbol string [``.\&?''] (arbitrary).

As illustrated in Figure \ref{fig:diff-delimiter}, our hypothesis is supported by the results. Structured delimiters consistently achieve the highest performance, which are the only delimiters that outperform baseline in all tasks. In contrast, semantic delimiters like ``and'' and ``seg'' often perform worse. This is presumably due to the semantic ambiguity they create, which force the model to disambiguate whether the token is a structural marker or part of the content. Arbitrary and natural delimiters show mixed results; while they outperform the baseline in some cases, the effect is inconsistent. 
It confirms that the performance gain does not stem from any specific semantic meaning, but rather from the introduction of a regular, discernible pattern. The advantage of structured tokens like ``<seg>'' resides in their function to provide a less ambiguous signal of sentence boundaries -- this enables the model to decouple structural processing from semantic reasoning.

\subsection{On the Optimality of Granularity: Sentence vs. Alternative Segmentations}

Having established the role of the delimiter's form, we now investigate its placement. Is segmentation at the sentence level inherently better than other granularities? We explore two alternatives: fixed-length chunking and random placement.

\begin{figure}[tbp]
    \centering
    \includegraphics[width=\linewidth]{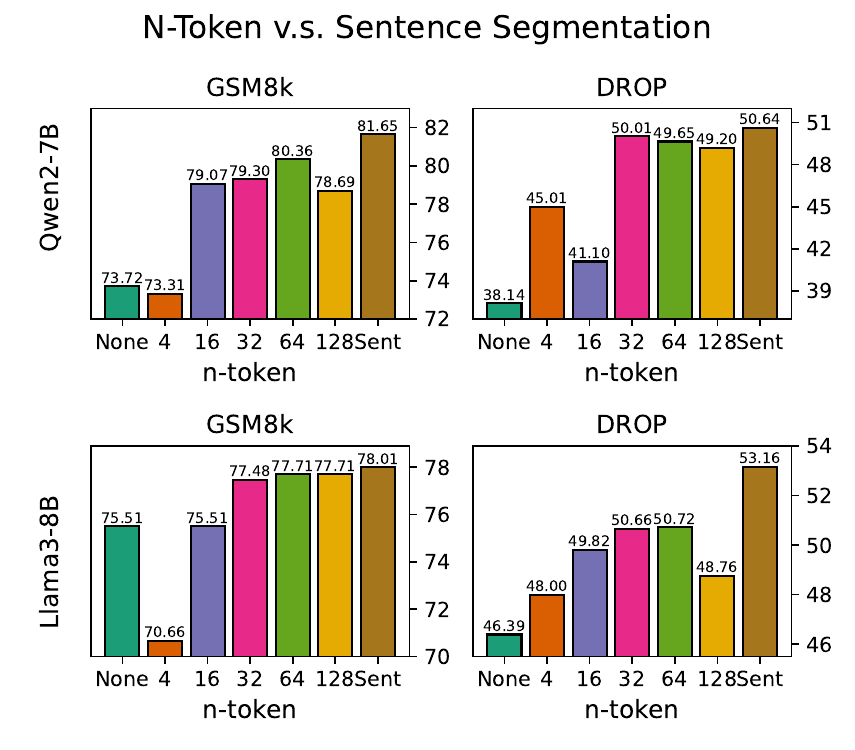}
    \caption{Sentence segmentation (Sent) vs. fixed n-token chunking. Sentence-level segmentation consistently outperforms fixed-chunking strategies, whose effectiveness decrease when the chunk size ($n$) is either too large or too small, only peaking when $n$ is close to the majority sentence length.}
    \label{fig:n-token-vs-sent}
\end{figure}

\paragraph{Comparison with Fixed-Length Chunking.} We replace sentence segmentation with a simple heuristic: inserting a delimiter every $n$ tokens. Figure \ref{fig:n-token-vs-sent} reveals a clear pattern: as $n$ increases, performance rises first, then falls. Very fine-grained chunking (e.g., $n=4,8$) is detrimental, as it fragments coherent semantic units within sentences. At the other end, very coarse-grained chunking (e.g., $n=128$) makes the structural signals too sparse to effectively guide step-by-step reasoning. The optimal performance is achieved within the range $n\in[32,64]$, which covers the typical sentence lengths in our test data (see \Cref{fig:dataset-info}). This strongly suggests that sentence is the ``natural'' unit of model reasoning: it balances between semantic integrity and the structural guidance function, which is a perfect analogy to how human process information, e.g., cognitive chunking\footnote{\url{https://dictionary.apa.org/chunking}}. 

\begin{figure}[tbp]
    \centering
    \includegraphics[width=0.9\linewidth]{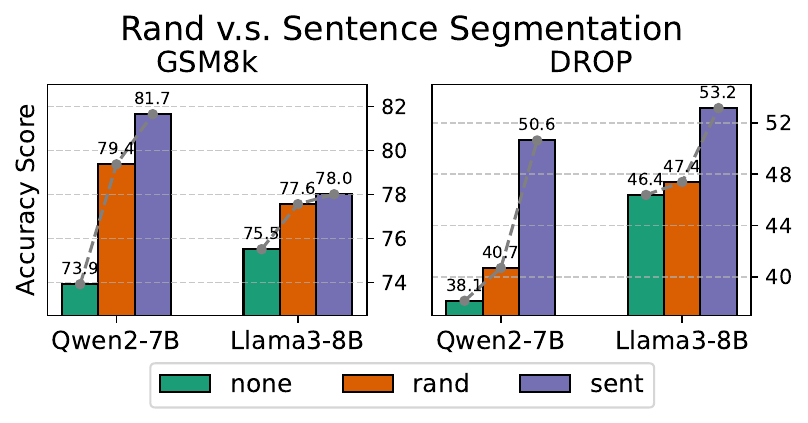}
    \caption{Sentence-level vs. random delimiter placement. Meaningful placement at sentence boundaries contributes more to the performance gains, far surpassing the minor effect of random insertions.}
    \label{fig:rand-vs-sent}
\end{figure}

\paragraph{Comparison with Random Placement.} To isolate the effect of delimiter positioning from the mere presence of additional tokens, we conducted a control experiment. For each input, we inserted the same number of delimiters as in sentence segmentation, but placed them at random positions. Results in Figure \ref{fig:rand-vs-sent} show that even random insertion yields a modest improvement over the baseline. This indicates what we term a minor ``dummy token'' effect: any regular interruption can slightly alter the model's processing. However, sentence-level placement consistently and significantly outperforms random placement. Therefore, we can conclude that the performance gain is not an artifact of adding extra tokens randomly, but is largely driven by placing delimiters at sentence boundaries -- positions that are meaningful and aligned with linguistic structure.

\subsection{Probing the Mechanism: Reasoning and Attention}

Why does sentence-level segmentation work so effectively? We investigate the mechanism from two perspectives: its role in the reasoning process and its effect on the model's attention patterns.

\paragraph{Enhancing Deliberative Reasoning.} We hypothesize that our method primarily benefits multi-step, deliberative reasoning rather than direct knowledge recall. To test this, we evaluate our fine-tuned model (Seg-FT and Std-FT) on MMLU using two zero-shot evaluation protocols: (1) Prob-based, which measures the model's immediate likelihood of the correct answer token, thereby probing knowledge recall; and (2) CoT-based, which prompts the model to generate a reasoning chain before the answer, hence probing deliberative reasoning.

\begin{table}[htbp]
    \centering
    \begin{tabular}{l|ccc}
    \toprule
         &  Std-FT & Seg-FT & Improvement \\
    \midrule
    Prob & 61.90 & 61.19 & -0.71\% \\
    CoT & 59.02 & 60.13 & +1.12\% \\
    \bottomrule
    \end{tabular}
    \caption{MMLU zero-shot performance of SFT models under two evaluation protocols. The benefits of our method manifest exclusively in the CoT setting, highlighting its role in enhancing deliberative reasoning.}
    \label{tab:prob-vs-cot}
\end{table}

Table \ref{tab:prob-vs-cot} shows a clear divergence. In the Prob-based setting, our method provides no benefit and even causes a slight degradation. However, in the CoT setting, it yields a clear improvement of +1.12\%. 
This result suggests that sentence-level delimiters do not simply improve the model's capabilities in retrieving static knowledge. Instead, the primary improvements are related to the dynamic, step-by-step reasoning process.

\paragraph{Attention as an Explanatory Lens.} To visualize the mechanism in terms of internal representations, we analyze the model's attention patterns. 
Examples of attention heatmaps (see \Cref{app:vis}) show that delimiter tokens act as focal points, drawing significant attention from subsequent tokens within the sequence. 

\begin{figure}[htbp]
    \centering    \includegraphics[width=0.9\linewidth]{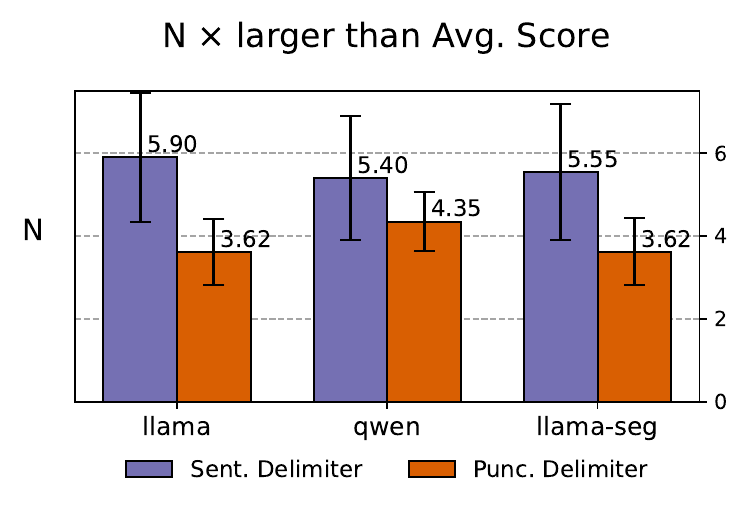}
    \caption{Relative attention scores for different delimiter types on the GSM8k dataset. Our delimiter (Sent. delimiter) receives significantly higher attention than both the sentence average (N$\times$ larger than avg.) and traditional punctuation delimiters (punc. delimiter).}
    \label{fig:n-larger-than-avg}
\end{figure}
For quantitative analysis, we compute the average attention paid to delimiter tokens by the final token of each sentence, and compare it against the attention paid to other tokens. As shown in Figure \ref{fig:n-larger-than-avg}, our special delimiter (sent. delimiter) receives substantially higher attention than other tokens on average. Interestingly, it attracts significantly more attention than natural punctuations (punc. delimiter) like periods or newlines. It indicates that the model has learned to treat the delimiter token as a more reliable ``signpost'' for demarcating the units of thought, compared to natural punctuation -- which is ambiguous and semantically overloaded. 
These delimiters thus function effectively as structural anchors, which the model can leverage to organize information flow during inference.

\section{Conclusions}

In this study we explore how explicitly modeling sentence structure in input can serve as a scaffold for enhancing the reasoning capabilities of Large Language Models in depth. We introduce a simple yet effective paradigm: teaching models to generate explicit boundary delimiters via in-context learning or fine-tuning. We validate the proposed methods through experiments on two directions: a lightweight, inference-time In-Context Learning strategy; and a more robust Supervised Fine-Tuning method that internalizes  prior knowledge on sentence structures directly into the model's parameters.

Our experiments are comprehensive in terms of model size, spanning from 7B to over 600B parameters, revealing consistent and significant performance gains across a diverse suite of reasoning benchmarks, including improvements of up to 7.7\% on GSM8k and 12.5\% on DROP. 
Our ablation studies further shed light on three key findings: (1) structurally distinct, non-semantic delimiters yield best effectiveness; (2) sentence is the optimal granularity for segmentation, outperforming both finer and coarser chunking strategies; and (3) the primary mechanism underlying the improvement is in facilitating of deliberative, step-by-step reasoning, a conclusion supported by both comparative analysis and attention visualization.

Beyond improving downstream task performance, our work also introduces a novel approach to structured text generation. By training LLMs to natively generate sentence-delimited output, we eliminate the computational overhead of post-hoc segmentation--a common requirement in applications like text-to-speech, retrieval-augmented generation, and controllable decoding. Therefore, this study validates a feasible pathway towards more efficient, structurally-aware, and capable language models, laying the ground for potential future explorations in cognitive-inspired LLM architectures.

Looking forward, we outline several promising research avenues for future research. Extending our SFT approach to the pre-training stage could potentially instill sentence awareness as a basic capability in foundation models. Furthermore, exploring the applicability of this method to low-resource languages and specialized domains (e.g., legal or medical texts) will be critical for assessing its universality. Finally, enabling models to perform self-segmentation has the potential to yield more adaptive and resource-efficient implementations.

\section{Limitations}

While our findings are promising, this study has several limitations that represent important directions for future work.

\paragraph{Generalization of Segmentation Methods.} Our experiments primarily rely on a state-of-the-art neural sentence segmenter (SaT). The robustness of our approach when using alternative segmentation methods, such as rule-based methods, or even the LLM's own self-segmentation capabilities, remains an open question. Investigating this is crucial for understanding the method's applicability in diverse, potentially resource-constrained production environments.

\paragraph{Validation at Larger Scales and Pre-training.} Although our ICL experiments include very large models, our supervised fine-tuning was conducted on 7B-level LLMs due to resource constraints. A full investigation of how sentence-aware fine-tuning interacts with scaling laws at a larger scale is a necessary next step. Furthermore, while our SFT results suggest strong potential, the ultimate impact of incorporating sentence-level objectives during the pre-training phase has yet to be empirically verified.

\paragraph{Deeper Interpretability.} Our analysis, based on attention scores and performance on reasoning-centric tasks, provides initial evidence for the mechanism behind our method's success. However, a more profound understanding is needed. Employing more advanced interpretability techniques, such as causal mediation analysis or probing for specific linguistic features in neuron activations, could more definitively trace how explicit structural signals modulate the model's internal computations and lead to improved reasoning.

\section*{Acknowledgments}
We sincerely thank all the reviewers for their feedback on the paper. This study is funded by Shenzhen Science and Technology Program (No. JCYJ20240813094612017) and Guangdong Province ZJRC Program (No. 2024QN11X145).

\bibliography{custom}

\onecolumn
\newpage
\twocolumn
\appendix

\section{Evaluation Settings} \label{app:fewshot}
Detailed evaluation settings of n-shot and CoT in ICL and SFT experiments are as follows:
\begin{itemize}[leftmargin=1em, itemsep=-0.1em]
    \item MMLU: 4-shot CoT for ICL, 0-shot CoT for SFT
    \item MMLU-Pro: 5-shot CoT for SFT
    \item GSM8k: 8-shot CoT for 7B-level LLMs and 4-shot CoT for large LLMs for ICL, 8-shot CoT for SFT
    \item MATH: 4-shot CoT for both ICL and SFT
    \item DROP: 3-shot for both ICL and SFT (DROP requires no CoT)
    \item GPQA: 0-shot CoT for SFT
    \item HumanEval: 0-shot for SFT (completion task cannot apply CoT)
\end{itemize}

\section{Combination between SFT with and without ICL}
In experiments, we assumed that the model obtained from sentence-segmented SFT would be used with ICL during inference on downstream tasks, which means the input of SFT model is well-segmented. This section explore whether a well-segmented input is strictly required by the SFT model.

\begin{table}[htbp]
    \centering
    \begin{tabular}{l|c|c}
    \toprule
         & GSM8k & DROP \\
    \midrule
    no-seg & 71.42 & 50.90 \\
seg	& 74.91	& 54.26 \\
\bottomrule
    \end{tabular}
    \caption{Comparison between SFT models with segmented input (seg) and raw input (no-seg).}
    \label{tab:seg_noseg}
\end{table}

Using the fine-tuned Llama3-8B model in \Cref{tab:sft_main}, we evaluated its performance on GSM8k and DROP under two conditions: with sentence-segmented input, and with raw, unsegmented input. As shown in \Cref{tab:seg_noseg}, the performance with segmented input is significantly better than without segmentation. This indicates that the SFT model has internalized the delimiter-augmented reasoning format; removing the delimiters leads to a distribution mismatch between training and evaluation, resulting in the performance degradation.

\section{Details about Sentence Segmentation Model} \label{app:sat}
All sentence segmentation is performed using \texttt{wtpsplist}, with default segmentation parameter \footnote{\url{https://github.com/segment-any-text/wtpsplit/blob/main/wtpsplit/__init__.py}}. Some details about the model \textsc{SAT-12L-sm}'s usage are listed below:
\begin{itemize}[leftmargin=1em, itemsep=-0.1em]
    \item stride: 256
    \item block\_size: 512
    \item pad\_last\_batch: False
    \item weighting: uniform
    \item model size: $\sim$300M
\end{itemize}

\section{SFT training details}
The SFT training parameters are listed below:

\begin{lstlisting}
trainer:
    use_flash_attn: true
    max_seq_length: 2048
    train_batch_size: 128
    learning_rate: 5.0e-06
    lr_scheduler_type: linear
    warmup_ratio: 0.03
    weight_decay: 0.0
    num_train_epochs: 1
deepspeed:
    zero_stage: 2
    gradient_clipping: 1.0
    offload: none
\end{lstlisting}


\newpage
\onecolumn

\section{An example of Segmented Input}

This is an example of segmented prompt and response from GSM8k to demonstrate how sentence-level inference works in our approaches. Delimiter here is ``<seg>''

\begin{tcolorbox}[title = {Example of delimited text}]
Mark has a garden with flowers. <seg> He planted plants of three different colors in it. <seg> Ten of them are yellow, and there are 80\% more of those in purple. <seg> There are only 25\% as many green flowers as there are yellow and purple flowers. <seg> How many flowers does Mark have in his garden? <seg> 
\tcblower
There are 80/100 * 10 = <\ <80/100 * 10=8>\ >8 more purple flowers than yellow flowers. <seg>
So in Mark's garden, there are 10 + 8 = <\ <10+8=18>\ >18 purple flowers. <seg>
Purple and yellow flowers sum up to 10 + 18 = <\ <10+18=28>\ >28 flowers. <seg>
That means in Mark's garden there are 25/100 * 28 = <\ <25/100 * 28=7>\ >7 green flowers. <seg>
So in total Mark has 28 + 7 = <\ <28+7=35>\ >35 plants in his garden. <seg>
\#\#\#\# 35 <seg>
\end{tcolorbox}

\section{Attention Map} \label{app:vis}

\begin{figure*}[h]
    \centering
    \includegraphics[width=0.9\linewidth]{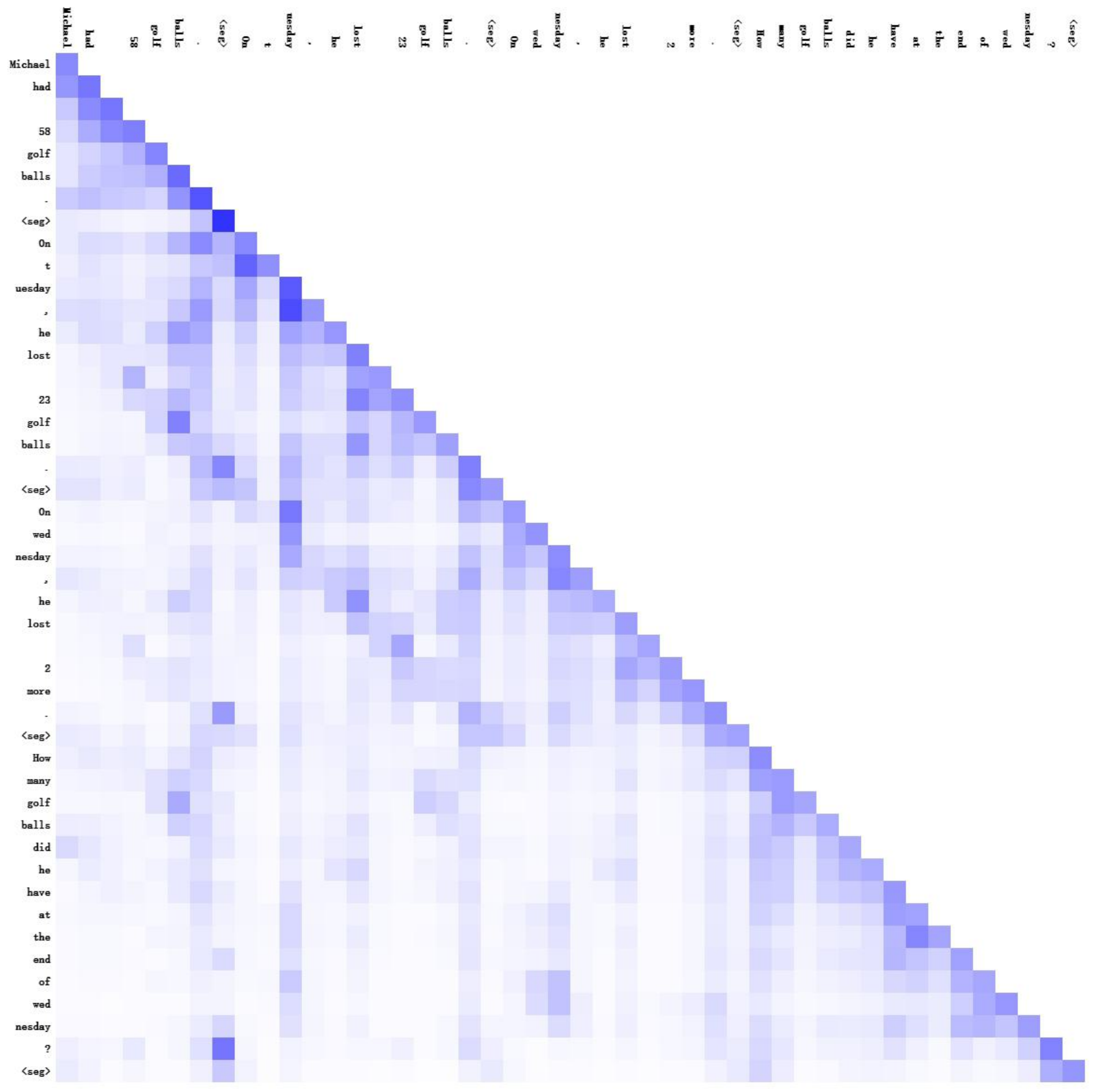}
    \caption{Attention map of Llama3-8b-seg}
    \label{fig:attnmapllamaseg}
\end{figure*}


\begin{figure*}[h]
    \centering
    \includegraphics[width=\linewidth]{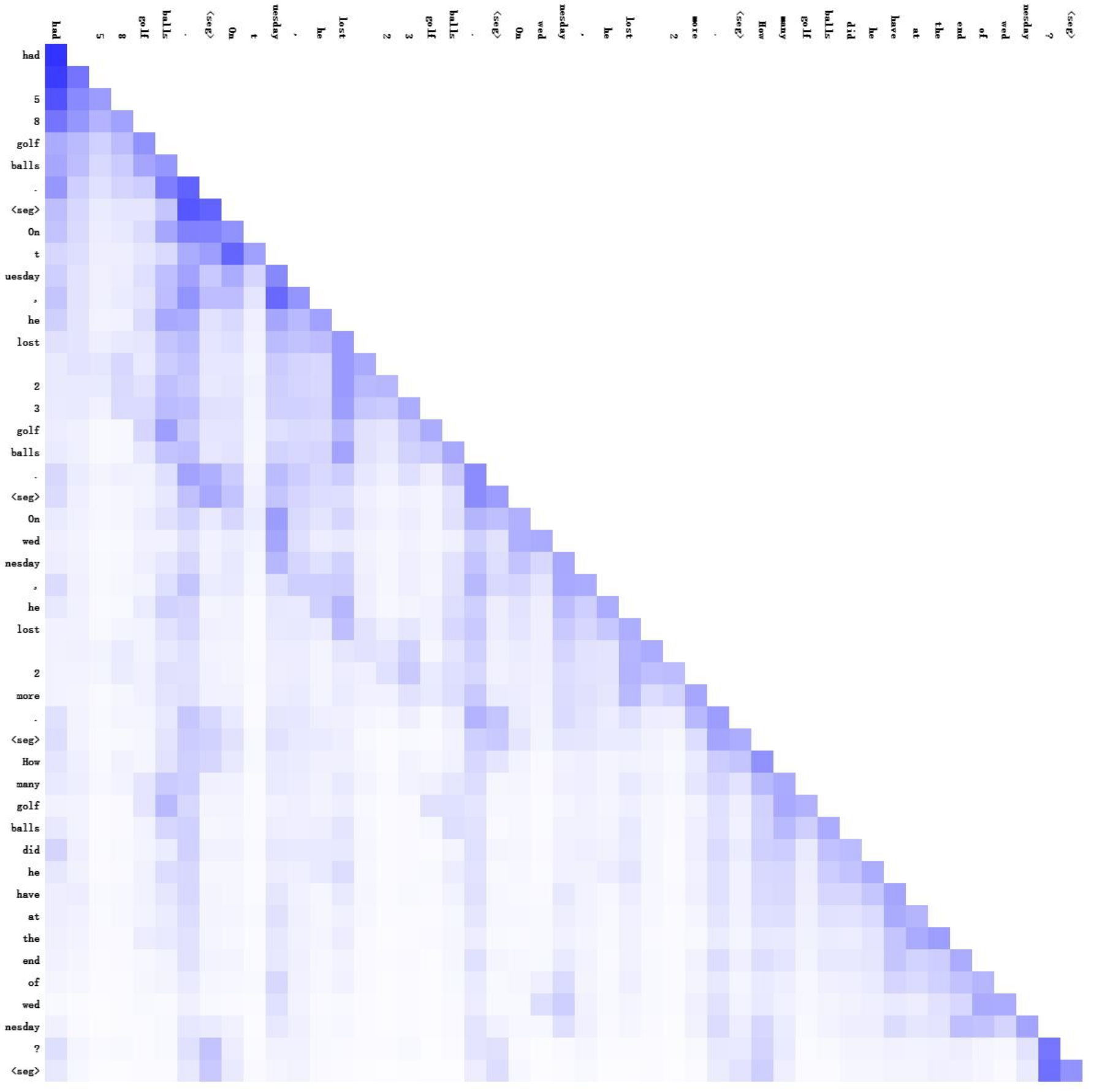}
    \caption{Attention map of Qwen2-7b-Instruct. The segmentation token we used is ``\textbf{\#\#\#\#''}. We replaced it to ``\textbf{\textless seg\textgreater}'' only when visualization.}
    \label{fig:attnmapqwen}
\end{figure*}

\end{document}